%% file: root.tex
\pdfoutput=1

\documentclass[letterpaper, 10pt, journal, twoside]{IEEEtran}

\IEEEoverridecommandlockouts    %

\usepackage[utf8]{inputenc}
\usepackage[T1]{fontenc}
\usepackage{amsmath}
\usepackage{amssymb}
\usepackage{amsfonts}
\usepackage{graphicx}
\usepackage{booktabs}
\usepackage{multirow}
\usepackage[table]{xcolor}
\usepackage{tikz}
\usepackage{siunitx}
\usepackage{soul}
\usepackage{url}
\usepackage{microtype}
\makeatletter
\let\NAT@parse\undefined
\makeatother
\usepackage[numbers,sort&compress]{natbib}
\usepackage[labelsep=colon,font=footnotesize,labelfont=bf,textfont=normalfont]{caption}
\usepackage[hidelinks]{hyperref}
\usepackage{eso-pic}

\title{IRIS: Inverse Dynamics from Random Exploration with Eye-in-hand Sensing}

\author{Xiaoxiong Zhang$^{1,*}$, Zhiqiang Zhou$^{1,*}$, Tze Ho Elden Tse$^{2}$, Linfang Zheng$^{3,\dagger}$, and Wei Zhang$^{1,4,\dagger}$%
\thanks{Manuscript received: August 25, 2026; Revised November 10, 2026; Accepted November 16, 2026.}%
\thanks{This paper was recommended for publication by Editor Wei Pan upon evaluation of the Associate Editor and Reviewers' comments. This work was supported by the Guangdong Science and Technology Program under Grant No. 2024B1212010002.}%
\thanks{$^{*}$Xiaoxiong Zhang and Zhiqiang Zhou contributed equally to this work. $^{\dagger}$Corresponding authors: Linfang Zheng and Wei Zhang.}%
\thanks{$^{1}$Xiaoxiong Zhang, Zhiqiang Zhou, and Wei Zhang are with Southern University of Science and Technology, Shenzhen, China\\{\tt\footnotesize \{12433017, 12532845\}@mail.sustech.edu.cn, zhangw3@sustech.edu.cn}}%
\thanks{$^{2}$Tze Ho Elden Tse is with the National University of Singapore, Singapore {\tt\footnotesize eldentse@comp.nus.edu.sg}}%
\thanks{$^{3}$Linfang Zheng is with The University of Hong Kong, Hong Kong {\tt\footnotesize lfzheng@hku.hk}}%
\thanks{$^{4}$Wei Zhang is also with LimX Dynamics, Shenzhen, China}%
\thanks{Digital Object Identifier (DOI): see top of this page.}}
\begin{document}
\AddToShipoutPictureBG*{\AtPageLowerLeft{\put(\LenToUnit{\dimexpr0.5\paperwidth-0.5\textwidth\relax},\LenToUnit{0.16in}){\parbox[b]{\textwidth}{\centering\scriptsize \copyright~2026 IEEE. Personal use of this material is permitted. Permission from IEEE must be obtained for all other uses, in any current or future media, including reprinting/republishing this material for advertising or promotional purposes, creating new collective works, for resale or redistribution to servers or lists, or reuse of any copyrighted component of this work in other works.}}}}

\bstctlcite{IEEEtran:BSTcontrol}

\maketitle

\input{sections/abstract}

\input{sections/intro}

\input{sections/related}

\input{sections/case_study}

\input{sections/method}

\input{sections/experiment}

\input{sections/limitation_future_work}

\input{sections/conclusion}

\bibliographystyle{IEEEtran}
\bibliography{refs}

\section*{APPENDIX}

\input{appendix/implementation}

\end{document}

%% file: sections/abstract.tex
\begin{abstract}
    Imagine-then-execute methods offer a promising paradigm for video-based manipulation, where a video planner predicts future visual subgoals and an inverse dynamics model (IDM) grounds them into robot actions. Despite their promise, these methods remain limited by action grounding. First, the IDM is typically trained on limited task-specific demonstrations and can generalize poorly even when given oracle visual subgoals, indicating that action grounding remains a bottleneck beyond visual planning errors.
Second, precise execution benefits from eye-in-hand observations, but eye-in-hand video planning is both hard to learn and hard to scale: the viewpoint moves with the robot, and such videos are scarce at internet scale. To overcome these challenges, we propose IRIS, a video-based manipulation framework that learns a highly generalizable, execution-aware IDM. First, IRIS scales action grounding by pretraining the IDM on task-agnostic random exploration data, which provides diverse visual state transitions paired with robot actions without human task annotation. Second, it extends the IDM with eye-in-hand sensing through a two-stage training scheme, without requiring eye-in-hand video planning. Extensive simulation and real-world experiments show that IRIS improves action grounding and data efficiency, enables more precise execution, and achieves competitive performance against state-of-the-art manipulation policies.
\end{abstract}

\begin{IEEEkeywords}
Imitation learning, machine learning for robot control, deep learning in grasping and manipulation.
\end{IEEEkeywords}

%% file: sections/intro.tex
\section{Introduction}
\label{sec:intro}

\IEEEPARstart{A}{} central goal of robot learning is to acquire manipulation policies that generalize across tasks, objects, and environments. Video-based policy learning offers a promising route toward this goal, since video models can be pretrained on large-scale video data and capture broad knowledge about visual dynamics and task structure~\cite{unipi,dreamzero,cosmos_policy}. A particularly attractive instantiation is the \emph{imagine-then-execute} paradigm, which decouples visual planning from robot control: a video planner first predicts future visual subgoals from the current observation and a language instruction, and an inverse dynamics model (IDM) then grounds these subgoals into robot actions~\cite{unipi,susie,atm,avdc,gvf-tape,clover}.
While this decomposition separates high-level visual planning from low-level action execution, successful manipulation requires both components to generalize. Recent work has substantially improved visual planners through larger datasets and stronger generative backbones, but the IDM has received comparatively less attention. This imbalance is problematic: even with a plausible visual plan, the robot succeeds only if the IDM can convert visual transitions into accurate actions. Scaling imagine-then-execute manipulation to general, fine-grained tasks therefore faces two closely related challenges.

\textbf{(i) Limited generalization of action grounding.} IDMs are typically trained on a small set of task-specific robot demonstrations~\cite{unipi,susie,atm,vidar}. These demonstrations are expensive to collect and cover only a narrow subset of possible object configurations, contact states, and robot motions. As a result, at test time the IDM may need to execute visual transitions that differ substantially from those seen during training. A natural hypothesis is that the resulting failures mainly come from imperfect visual subgoals produced by the video planner. In a controlled pilot study (Section~\ref{sec:pilot}), we replace generated subgoal images with \emph{oracle} frames from successful demonstrations to eliminate image-generation errors, while varying initial configurations between training and evaluation. Even in this favorable setting, the IDM loses $57.2$ pp of success rate on held-out configurations, indicating that \emph{action grounding itself is a key generalization bottleneck}: a perfect visual plan does not guarantee successful execution when the IDM is trained on narrow demonstration data.

\textbf{(ii) Difficulty of integrating eye-in-hand sensing.}
Eye-in-hand observations provide local feedback for occlusion, fine alignment, and contact, and are widely used by manipulation policies~\cite{diffusion_policy,act,Openvla-oft}.
Incorporating this signal into imagine-then-execute systems, however, is less straightforward. A direct extension would require the video planner to generate future eye-in-hand subgoals alongside side-view ones. This is challenging for two reasons: eye-in-hand views are tightly coupled to robot motion, making them inherently harder to predict than stable side-view scenes; and eye-in-hand videos are much less available than third-person videos at internet scale, undermining the data scalability that motivates video-based planning in the first place. As a result, most imagine-then-execute systems plan only in the side view~\cite{unipi,susie,atm,gvf-tape,clover}. This raises a key research question: can an IDM benefit from eye-in-hand observations during execution without requiring the video planner to generate eye-in-hand subgoals?

We address both challenges with a single framework that intervenes entirely at the IDM: it changes what data the IDM is trained on and how it is conditioned, while leaving the side-view video planner intact. Our key observation is that action-grounding supervision---tuples of current observations, future observations, and executed actions---does not require task labels. We can obtain it from \emph{random exploration data}, where the robot executes randomly sampled actions without human supervision. Although task-agnostic, such data exposes the robot to a broader range of motions and visual transitions than narrow task demonstrations. We use it to pretrain the IDM to follow visual subgoals before task-specific adaptation, directly addressing the action-grounding generalization gap.

Random exploration also provides a natural way to incorporate eye-in-hand sensing. Because the eye-in-hand camera moves with the robot during exploration, the same trajectories contain both side-view and eye-in-hand observations \emph{for free}. We exploit this signal with a two-stage IDM training scheme: during random-exploration pretraining, the IDM is conditioned on \emph{both} side-view and eye-in-hand future subgoals, allowing it to learn how actions affect both scene-level motion and local gripper-centric changes; during task-specific fine-tuning, we mask the future eye-in-hand subgoal, letting the IDM benefit from eye-in-hand sensing at execution while keeping visual planning side-view-only and scalable. The resulting IDM uses the current eye-in-hand observation as live execution context, but the system never needs to imagine future eye-in-hand frames.

We integrate these ideas into \textbf{IRIS}, a video-based manipulation framework that pairs a side-view RGB-D video planner with an \underline{I}nverse-dynamics model trained from \underline{R}andom exploration and \underline{I}n-hand \underline{S}ensing. 
Our contributions are:

\begin{itemize}
    \item \textbf{A scalable route to stronger action grounding.} We show that task-agnostic random exploration can serve as a scalable pretraining source for IDMs, reducing their dependence on narrow task demonstrations and improving generalization to held-out configurations. 

    \item \textbf{Eye-in-hand sensing without sacrificing video-planning scalability.} We introduce a two-stage training scheme that lets the IDM exploit eye-in-hand observations for precise manipulation while keeping visual planning entirely in the side-view space, avoiding the need to generate eye-in-hand video subgoals.

    \item \textbf{A validated framework for video-based manipulation.} We instantiate these ideas in IRIS; controlled studies and comparisons with strong baselines show gains in action-grounding generalization, data efficiency, and execution precision.
\end{itemize}

%% file: sections/related.tex
\section{Related Work}
\label{sec:related}

\textbf{Imagine-then-execute Manipulation.}
A growing line of work casts robot manipulation as video-based policy learning, differing in how visual prediction is coupled to control: jointly modeling future videos and actions~\cite{dreamzero,cosmos_policy,gr2}, using video prediction as an auxiliary training objective~\cite{fast-wam,uwm,uva,univla,motus}, and extracting representations from pretrained video models for downstream policies~\cite{video-policy,video-prediction-policy,mimic-video,dit4dit}. Most relevant to IRIS is the \emph{imagine-then-execute} paradigm, which decouples a video planner that generates future visual subgoals from an inverse dynamics model (IDM) that grounds them into actions~\cite{unipi,amplify,atm,avdc,dreamitate,lingbot-va,susie,vidar,gvf-tape,v2a,clover}. Many of these methods focus on improving the planner, including generative quality, language conditioning, and temporal consistency. We instead identify the IDM as an equally important but underexplored bottleneck, and focus our contributions there.

\textbf{Scaling Action Grounding beyond Demonstrations.}
IDMs in imagine-then-execute systems commonly learn from task-specific demonstrations~\cite{unipi,susie,atm,vidar}, which cover a narrow range of object configurations and contact states. Vidar~\cite{vidar} improves IDM generalization through implicit mask prediction, while V2A~\cite{v2a} uses online exploration guided by video-generated goals, making its data collection task-directed. Random exploration has also supported visual world modeling~\cite{world-models,visual-foresight,gpc} and self-supervised grasping~\cite{RoboNet,qt-opt,hand-eye-coordinating-grasp,superrizing-self-supervise}, but its use for offline visual-subgoal IDM pretraining remains limited. GVF-TAPE~\cite{gvf-tape} uses random exploration to estimate poses from visual plans and follows them with a pose-tracking controller. IRIS instead pretrains an IDM on offline, task-agnostic random transitions and predicts action chunks directly from scene and wrist observations.

\textbf{Eye-in-hand Sensing in Manipulation Policies.}
Close-range feedback from a wrist camera captures cues---occluded contacts and fine alignment near the gripper---that side-view cameras can miss, and is a common input to imitation-learning and vision--language--action policies~\cite{diffusion_policy,act,pi_0,pi_05,Openvla-oft}. Imagine-then-execute manipulation, however, introduces a design tension: most systems plan only in a fixed side view~\cite{unipi,susie,avdc,gvf-tape,clover}, while recent works such as LingBot-VA and Vidar~\cite{lingbot-va,vidar} extend the planner to generate future wrist views as well. The latter inherits the difficulties of egocentric video prediction, since the viewpoint moves with the end effector and large wrist-view corpora are scarce. This leaves open how an IDM can exploit eye-in-hand observations without requiring the planner to imagine them. IRIS uses future wrist frames during IDM pretraining, masks them during task adaptation, and retains the current wrist image for execution.

%% file: sections/case_study.tex
\section{A Pilot Study: Isolating the Action-Grounding Bottleneck}
\label{sec:pilot}

\input{figs/pilot_studying}

\input{figs/framework}

To examine whether action grounding is a real generalization bottleneck or merely a downstream symptom of imperfect visual planning, we conduct a controlled experiment on LIBERO-Long~\cite{libero}, which contains 10 tasks with 50 demonstrations per task. We replace the video planner with \emph{oracle} visual subgoals---future frames from the corresponding expert demonstration---and vary the initial scene configuration between training and evaluation. This setup removes image-generation error and tests the IDM's ability to ground unfamiliar visual transitions.

We train a diffusion-based IDM without eye-in-hand observations on the first 30 demonstrations per task and evaluate on all 50, reporting success separately on the 30 seen demonstrations and the 20 held-out ones with unseen initial configurations. As shown in Fig.~\ref{fig:pilot}(a), the IDM reaches $76.7\%$ on the seen demonstrations but only $19.5\%$ on the held-out ones---a $57.2$ percentage-point gap despite oracle visual subgoals. This shows that action grounding itself can fail to generalize to unseen scene configurations.

Task demonstrations expose the IDM to only a narrow slice of object configurations, contact states, and robot motions. To test whether this limited coverage contributes, we pretrain the same IDM on task-agnostic random exploration and fine-tune on them. Pretraining lifts held-out success from $19.5\%$ to $56.5\%$ and seen success from $76.7\%$ to $87.7\%$.

Oracle frames are time-indexed and consumed open-loop, so they can misalign with the robot once execution deviates. Figure~\ref{fig:pilot}(b) therefore repeats the evaluation with subgoals generated closed-loop from the current observation by the video planner of Section~\ref{sec:experiment}. Both settings share IDM training data and initial configurations; the generated-subgoal planner sees all 50 videos per task, including those held out from IDM training. Closed-loop generated subgoals raise held-out success without pretraining from $19.5\%$ to $38.0\%$, but a $29.3$ pp seen--held-out gap remains. Pretraining raises generated-subgoal success to $73.0\%$ and narrows that gap to $13.3$ pp.

As shown in Fig.~\ref{fig:pilot}(c), pretraining also reduces the deviation between executed and oracle trajectories on Task 2 throughout execution. Together, these results establish that the action-grounding generalization gap is real, present under both oracle and generated subgoals, and substantially reduced through broader pretraining---motivating random exploration as a scalable pretraining source for IRIS.

%% file: figs/pilot_studying.tex
\begin{figure}[t]
    \centering
    \includegraphics[width=0.825\columnwidth]{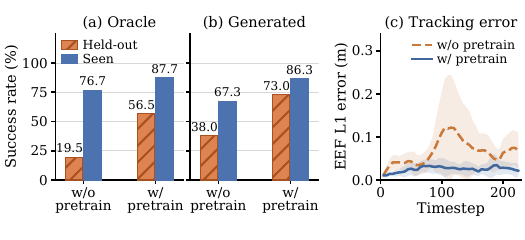}
    \vspace{-5mm}
    \caption{\footnotesize \textbf{Pilot study: action grounding fails to generalize, even with oracle visual subgoals.}
    \textbf{(a)} Success rate over 10 LIBERO-Long tasks on seen versus held-out initial configurations, with and without random-exploration pretraining.
    \textbf{(b)} As (a), but with subgoals generated by the video planner instead of oracle frames.
    \textbf{(c)} Mean L1 $(x,y,z)$ error between executed and oracle end-effector trajectories on Task~2 over 20 held-out trials; shading denotes standard deviation.}
    \label{fig:pilot}
\end{figure}

%% file: figs/framework.tex
\begin{figure*}[t]
    \vspace*{0.25mm}
    \centering
    \includegraphics[width=0.7\textwidth]{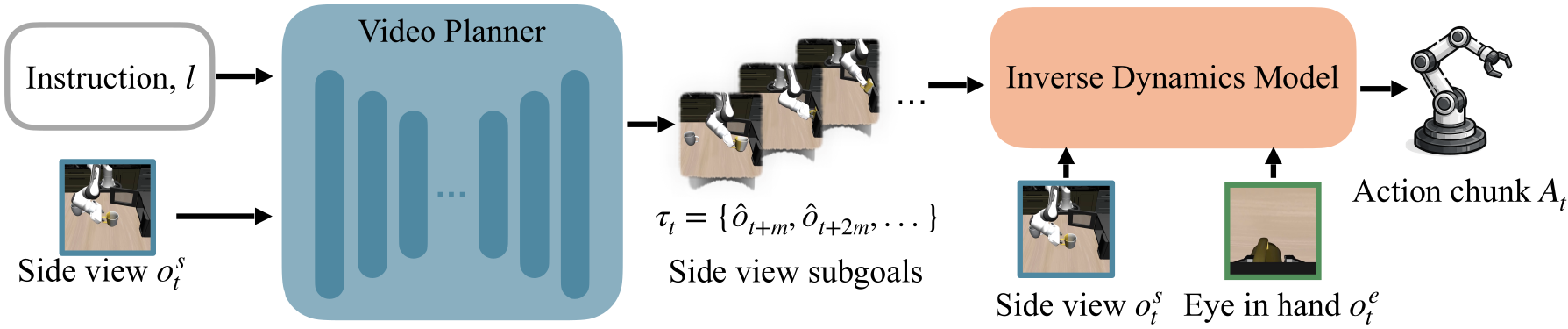}
    \vspace{-1mm}
    \caption{\footnotesize \textbf{Framework overview.}
    A side-view video planner predicts a sequence of future visual subgoals $\tau_t$ from the current observation $o_t^s$ and language instruction $l$. The IDM grounds each subgoal $\hat{o}_{t+m}^s$ into an action chunk $A_t$, conditioned on the current side-view $o_t^s$ and eye-in-hand $o_t^e$ observations. The robot replans in closed loop after each visual plan is consumed. }
    \label{fig:framework}
    \vspace{-2mm}
\end{figure*}

%% file: sections/method.tex
\section{Method}

IRIS follows the imagine-then-execute paradigm, but shifts the main learning burden from generating additional visual plans to training a more capable execution module. As shown in Fig.~\ref{fig:framework}, a side-view video planner (Section~\ref{subsec:videogen}) supplies visual subgoals to an inverse dynamics model (IDM) that grounds them into action chunks. Both challenges identified in the introduction are addressed at the IDM via a two-stage training scheme (Section~\ref{subsec:idm_training}, Fig.~\ref{fig:idm}): random-exploration pretraining improves action-grounding generalization, and eye-in-hand-masked fine-tuning integrates eye-in-hand sensing without generating eye-in-hand subgoals.

\subsection{Framework Overview}
\label{subsec:overview}

At each planning step $t$, IRIS receives a side-view image \(o_t^s\), an eye-in-hand image \(o_t^e\), and a language instruction \(l\), and produces an \(m\)-step action chunk \(A_t = \{a_t, \ldots, a_{t+m-1}\}\), where \(m\) matches the temporal stride of the visual subgoals. The side-view image provides a global view of the scene for planning, while the eye-in-hand image provides local visual feedback for execution.

The video planner \(p_\theta\) samples an \(H\)-step side-view visual plan from the current observation,
\vspace{-1mm}
\begin{equation}
\vspace{-1mm}
 \tau_t = \{\hat{o}_{t+m}^s, \hat{o}_{t+2m}^s, \ldots, \hat{o}_{t+Hm}^s\}
\sim p_\theta(\cdot \mid o_t^s, l),   
\end{equation}
which the robot then consumes one subgoal at a time. For each upcoming subgoal \(\hat{o}_{t+m}^s\), the IDM \(\pi_\phi\) generates the next \(m\)-step action chunk conditioned on the current side-view and eye-in-hand observations,
\vspace{-1mm}
\begin{equation}
\vspace{-1mm}
    A_t \sim \pi_\phi\!\left( \cdot \mid o_t^s, o_t^e, \hat{o}_{t+m}^s \right),
\end{equation}
where \(t\) advances by \(m\) after each chunk is executed. Once all \(H\) subgoals have been consumed, the planner re-samples from the latest observation, closing the loop.

\subsection{Side-view Video Planning}
\label{subsec:videogen}

We plan only in the side-view space because side-view videos are more stable under robot motion and more compatible with scalable video pretraining than eye-in-hand videos; Section~\ref{subsec:real_robot_eih} compares this choice against a joint dual-view generation alternative. To provide spatial cues for action grounding, we generate \emph{RGB-D} plans from RGB observations~\cite{clover,gvf-tape}: the depth channel is supervised by pseudo-depth targets from an off-the-shelf video depth estimator~\cite{videodepth}, keeping the training pipeline compatible with RGB-only video data while avoiding a depth sensor at inference. For brevity, \(\tau_t\) denotes this RGB-D plan in the remainder of the section.

We instantiate \(p_\theta(\tau_t \mid o_t^s, l)\) with rectified flow~\cite{rflow}. Given a clean plan \(\tau_t^1 \sim \mathcal{D}\), Gaussian noise \(\tau_t^0 \sim \mathcal{N}(0, I)\), and a flow time \(r \in [0,1]\), we form the linear interpolant \(\tau_t^r = (1-r)\tau_t^0 + r\tau_t^1\) and learn a conditional velocity field \(v_\theta(\tau_t^r, r, o_t^s, l)\) that transports noise to visual plans by minimizing the flow-matching objective
\[
\mathcal{L}
=
\mathbb{E}_{\substack{\tau_t^1 \sim \mathcal{D},\, \tau_t^0 \sim \mathcal{N}(0,I),\\ r \sim \mathcal{U}(0,1)}}
\!\left[
\big\| v_\theta(\tau_t^r, r, o_t^s, l) - (\tau_t^1 - \tau_t^0) \big\|_2^2
\right].
\]
We parameterize \(v_\theta\) with a lightweight 3D U-Net~\cite{avdc} and encode \(l\) with a CLIP text encoder~\cite{CLIP}. At inference, we sample \(\tau_t^0 \sim \mathcal{N}(0, I)\) and integrate \(d\tau_t^r/dr = v_\theta(\tau_t^r, r, o_t^s, l)\) from \(r{=}0\) to \(r{=}1\) to obtain the RGB-D plan. 

\subsection{Two-Stage IDM Training}
\label{subsec:idm_training}

The IDM is responsible for action grounding: given the current observations and a future visual subgoal, it predicts the action chunk that moves the robot toward that subgoal. {We instantiate the IDM with Diffusion Policy~\cite{diffusion_policy}, a probabilistic action-chunk model that can represent multimodal action distributions for the same visual transition. Such multimodality is important because multiple short-horizon motions may achieve similar visual progress, especially under contact or partial observability.} We train the IDM in two stages, illustrated in Fig.~\ref{fig:idm}.

\paragraph{Stage 1: Random-exploration pretraining.}
Training the IDM only on task demonstrations gives limited coverage of robot motions and scene configurations. IRIS therefore pretrains the IDM on \emph{random exploration data}, collected by executing randomly sampled actions without task labels. This data is task-agnostic, yet directly supervises inverse dynamics by pairing visual state transitions with the actions that caused them.

During pretraining, each sample contains current observations \((o_t^s, o_t^e)\), future observations \((o_{t+m}^s, o_{t+m}^e)\), and the executed action chunk \(A_t\) from a random-exploration dataset \(\mathcal{D}_{\mathrm{rand}}\). We train the IDM by maximizing the likelihood of the executed action chunk:
\[
\max_{\phi}\;
\mathbb{E}_{\substack{(o_t^s, o_t^e, o_{t+m}^s, o_{t+m}^e, A_t)\\ \sim\, \mathcal{D}_{\mathrm{rand}}}}
\!\left[
\log \pi_\phi\!\left( A_t \mid o_t^s, o_t^e, o_{t+m}^s, o_{t+m}^e \right)
\right].
\]
The side-view future subgoal \(o_{t+m}^s\) provides scene-level supervision for action grounding, while the eye-in-hand future subgoal \(o_{t+m}^e\) provides local gripper-centric supervision. This dual-view supervision helps the IDM acquire a general visual-subgoal-following ability before 
seeing task-specific demonstrations.

{\paragraph{Stage 2: Task adaptation without eye-in-hand planning.}
By design, IRIS confines visual planning to the side-view space for scalability. At the same time, we require the IDM to use eye-in-hand sensing for precise execution. Stage 2 reconciles these two design goals: starting from the pretrained IDM, we fine-tune on task demonstrations \(\mathcal{D}_{\mathrm{task}}\) while masking the future eye-in-hand subgoal as \(\varnothing\),
\[
\max_{\phi}\;
\mathbb{E}_{(o_t^s, o_t^e, o_{t+m}^s, A_t) \sim \mathcal{D}_{\mathrm{task}}}
\!\left[
\log \pi_\phi\!\left( A_t \mid o_t^s, o_t^e, o_{t+m}^s, \varnothing \right)
\right].
\]}

{At deployment, the generated side-view subgoal \(\hat{o}_{t+m}^s\) replaces the demonstration subgoal \(o_{t+m}^s\), and the IDM predicts
\begin{equation}
A_t \sim \pi_\phi\!\left( \cdot \mid o_t^s, o_t^e, \hat{o}_{t+m}^s, \varnothing \right). 
\end{equation}
Thus, the IDM uses the current eye-in-hand image as live execution feedback, but IRIS never asks the video planner to predict future eye-in-hand frames. This design preserves the scalability of side-view video planning while retaining the local sensing benefits needed for precise manipulation.}

%% file: sections/experiment.tex
\section{Experiments}
\label{sec:experiment}

{We evaluate whether the two IDM-side interventions of IRIS---random-exploration pretraining and eye-in-hand conditioning---relieve the action-grounding bottleneck identified in Section~\ref{sec:pilot}. We first compare the full system against state-of-the-art policies on LIBERO, then isolate what each intervention contributes in simulation and on a real robot.}

\begin{figure}[t]
    \centering
    \includegraphics[width=0.48\textwidth]{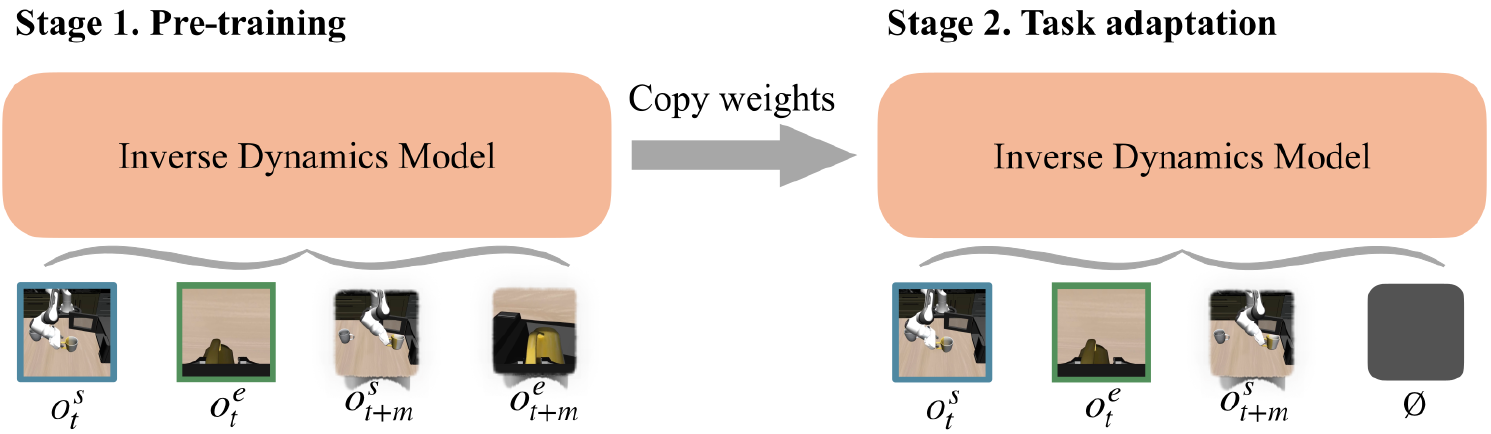}
    \vspace{-1mm}
    \caption{\footnotesize \textbf{Two-stage IDM training.}
    \textbf{Stage 1} pretrains on random exploration data with both side-view and eye-in-hand future subgoals \((o_{t+m}^s, o_{t+m}^e)\). \textbf{Stage 2} fine-tunes on task demonstrations with the eye-in-hand subgoal replaced by a null placeholder \(\varnothing\), matching the side-view-only inference setting.
    }
    \label{fig:idm}
\end{figure}

\input{figs/realworld_setup}

\subsection{Experimental Setup}
\label{subsec:exp_setup}

\textbf{Simulation setup.}
{We use the LIBERO benchmark~\cite{libero}, which contains four suites---LIBERO-Spatial, LIBERO-Object, LIBERO-Goal, and LIBERO-Long---covering spatial variation, object variation, goal variation, and long-horizon manipulation, respectively. Each suite contains $10$ tasks with $50$ expert demonstrations per task. We evaluate each method over $50$ trials per task and report success rate. Unless noted otherwise, IRIS results with $\pm$ or error bars are means over three IDM fine-tuning seeds, with the planner and exploration pretraining fixed, and the spread is one standard deviation; those without are single fine-tuning runs.}

{We use two complementary data sources in simulation. \emph{Task-specific demonstrations} train the video planner and fine-tune the IDM, while \emph{task-agnostic random-exploration} transitions pretrain the IDM. We collect $240$K random-exploration steps each for LIBERO-Spatial, LIBERO-Object, and LIBERO-Goal, and $1.2$M steps for LIBERO-Long given its longer horizons. The exploration set is generated by executing random robot actions, without task labels, language instructions, or success annotations; each pretraining sample pairs two observations with the actions executed between them. In total, exploration contributes $1.92$M transitions versus $339$K demonstration transitions. {Pretraining on the exploration data is a one-time cost of about $4$ hours on four A100 GPUs for the three short-horizon suites ($18$ hours for LIBERO-Long).} {In experiments that vary the demonstration budget, we reduce only the action-labeled demonstrations used to fine-tune the IDM, while keeping the video planner trained on the full demonstration videos, following prior work~\cite{atm,ssi-policy}.}  Full architecture and training details for the video planner and IDM are provided in the Appendix.}

\textbf{Real-world setup.}
{We deploy IRIS on an ARX-5 robotic arm with a side-view RealSense D435i camera and a wrist camera, and evaluate the five tasks in Fig.~\ref{fig:real_world_setup}, which stress complementary execution challenges: \emph{(T1)} inserting a block into a slot, which requires precise alignment; \emph{(T2)} picking a lemon and placing it on a plate that is partly occluded from the side view; \emph{(T3)} placing a lemon into a cabinet drawer and closing it; \emph{(T4)} closing a microwave door; and \emph{(T5)} placing a red mug on a mug rack.} For each task, we collect $50$ teleoperated demonstrations, plus $23$K task-agnostic random-exploration steps. The video planner is trained on all $250$ demonstrations across the five tasks, and the IDM is fine-tuned on the same data; each method is evaluated over $30$ trials per task. {Inference runs on a single NVIDIA RTX 4090 GPU with FP16 quantization, taking $0.350$\,s per video-planner call and $0.139$\,s per IDM action prediction.}

\textbf{Baselines.}
{We compare IRIS against representative methods spanning imitation learning~\cite{diffusion_policy,mdt}, video-based policy learning~\cite{mimic-video,ssi-policy,seer,cosmos_policy,lingbot-va}, and vision--language--action models~\cite{Openvla,Openvla-oft,cotvla,clip-rt,pi_0,pi_05,univla}. Because external pretraining strongly affects performance, we group baselines by whether they are initialized from pretrained robot policies, vision--language--action models, or video backbones; {``Scratch'' variants, marked $\dagger$ in Table~\ref{tab:main_results}, train the same architecture without such initialization. IRIS belongs to the first group: it uses a frozen CLIP text encoder~\cite{CLIP} for language conditioning but no pretrained policy or video backbone. {IRIS also uses a pretrained video depth estimator~\cite{videodepth} to produce pseudo-depth training targets, but does not use it at inference.}} On the real robot, we additionally compare against Diffusion Policy and GVF-TAPE~\cite{gvf-tape} trained on the same demonstrations, and against two IRIS variants that remove eye-in-hand conditioning or exploration pretraining.}

\subsection{Comparison with State-of-the-Art on LIBERO}
\label{subsec:sota}

{We compare IRIS with prior methods on LIBERO under both the full demonstration budget (Table~\ref{tab:main_results}) and a low-data setting using $20\%$ of the demonstrations (Table~\ref{tab:data_efficiency}), evaluating overall performance and data efficiency, respectively. IRIS additionally pretrains its IDM on the task-agnostic random exploration described above, {which carries no task or language labels}, while the baselines follow their published training protocols. Section~\ref{subsec:component_ablation} isolates the contribution of exploration pretraining under matched demonstration budgets.}

\begin{table}[t]
\vspace*{0.25mm}
\centering
\caption{\footnotesize \textbf{Success rates (\%) on the LIBERO benchmark and policy size.} \textbf{Bold}/\underline{underlined}: column best/second-best within each group. $^\dagger$: \emph{Scratch} variant, trained without 
{a pretrained planner or policy backbone}. NR: not reported; --: unavailable.}
\label{tab:main_results}
\vspace{-1mm}
\scriptsize
\setlength{\tabcolsep}{3pt}
\renewcommand{\arraystretch}{1.1}
\begin{tabular*}{\columnwidth}{@{\extracolsep{\fill}}lcccccr@{}}
\toprule
\textbf{Method} & \textbf{Spatial} & \textbf{Object} & \textbf{Goal} & \textbf{Long} & \textbf{Avg.} & \textbf{Params} \\
\midrule
\multicolumn{7}{l}{\textit{{W/o pretrained planner/policy backbone}}} \\
\midrule
Diffusion Policy~\cite{diffusion_policy}     & 78.3 & 92.5 & 68.3 & 50.5 & 72.4 & $\approx$278M \\
MDT~\cite{mdt}                               & 78.5 & 87.5 & 73.5 & 64.8 & 76.1 & $\approx$375M \\
mimic-video~\cite{mimic-video}               & 94.2 & 96.8 & 90.6 & --   & --   & $>$2B \\
SSI-Policy~\cite{ssi-policy}                 & 94.0 & 97.5 & 93.0 & 80.5 & 91.3 & NR \\
OpenVLA-OFT$^\dagger$~\cite{Openvla-oft}       & \underline{94.3} & 95.2 & 91.7 & 86.5 & 91.9 & $\approx$7.71B \\
Cosmos Policy$^\dagger$~\cite{cosmos_policy}   & \textbf{94.7} & \textbf{98.9} & \underline{96.3} & \underline{88.6} & \underline{94.6} & $\approx$2B \\
\rowcolor{gray!10}[0pt][0pt]
\textbf{IRIS (Ours)}                         & 90.8 & \underline{98.0} & \textbf{96.6} & \textbf{96.3} & \textbf{95.4} & $\approx$551M \\
\rowcolor{gray!10}[0pt][0pt]
\multicolumn{1}{r}{\tiny\color{gray}std over seeds} & {\tiny\color{gray}$\pm$0.5} & {\tiny\color{gray}$\pm$0.3} & {\tiny\color{gray}$\pm$0.7} & {\tiny\color{gray}$\pm$1.6} & {\tiny\color{gray}$\pm$0.5} & \\
\midrule
\multicolumn{7}{l}{\textit{{W/ pretrained planner/policy backbone}}} \\
\midrule
OpenVLA~\cite{Openvla}                       & 84.7 & 88.4 & 79.2 & 53.7 & 76.5 & $\approx$7.54B \\
CoT-VLA~\cite{cotvla}                        & 87.5 & 91.6 & 87.6 & 69.0 & 83.9 & $\approx$7B \\
CLIP-RT+~\cite{clip-rt}                      & 95.2 & 99.2 & 94.2 & 82.6 & 92.8 & $\approx$1.3B \\
$\pi_0$~\cite{pi_0}                          & 96.8 & 98.8 & 95.8 & 85.2 & 94.2 & $\approx$3.3B \\
Seer~\cite{seer}                             & --   & --   & --   & 87.7 & --   & $\approx$316M \\
$\pi_{0.5}$~\cite{pi_05}                     & \textbf{98.8} & 98.2 & \underline{98.0} & 92.4 & \underline{96.9} & $\approx$3.3B \\
UniVLA~\cite{univla}                         & 96.5 & 96.8 & 95.6 & 92.0 & 95.2 & $\approx$7.55B \\
Cosmos Policy~\cite{cosmos_policy}           & 98.1 & \textbf{100.0} & \textbf{98.2} & \underline{97.6} & \textbf{98.5} & $\approx$2B \\
LingBot-VA~\cite{lingbot-va}                 & \underline{98.5} & \underline{99.6} & 97.2 & \textbf{98.5} & \textbf{98.5} & $>$5B \\
\bottomrule
\end{tabular*}
\vspace{-0.25mm}
\end{table}

{{\textbf{Full-demonstration comparison.} Among methods {without a pretrained planner or policy backbone}, IRIS attains the highest average success rate while using roughly a quarter of the parameters of Cosmos Policy (Scratch), the next-best method. Its largest advantage is on LIBERO-Long, where it leads by $7.7$ percentage points (pp): sustained multi-stage execution makes accumulated action-grounding errors especially consequential. IRIS is also within $0.9$ pp of the best Object result and {exceeds the next-best Goal result by $0.3$ pp} ($96.6$ vs.\ $96.3$). Its weaker Spatial result is concentrated in one task ($31\%$ success averaged over seeds; $97.4\%$ over the remaining nine), where the planner generates implausible subgoals---the bowl rises before the fingers close---that the IDM then follows.}

{Across all methods with reported averages, IRIS ranks fourth overall {with only $\approx551$M parameters; the three methods ahead of it all rely on pretrained backbones and substantially larger models}. Cosmos Policy, the only method reported in both variants, indicates the size of that advantage: external initialization lifts its average by $3.9$ pp. {Strengthening action grounding can thus be competitive without large pretrained policies or video backbones.}}

{\textbf{Limited-demonstration comparison.}
{With $20\%$ of the {action-labeled} demonstrations ($10$ per task), IRIS achieves the best result on LIBERO-Object and LIBERO-Long and the second-best on LIBERO-Spatial and LIBERO-Goal (Table~\ref{tab:data_efficiency}). {GVF-TAPE~\cite{gvf-tape}, the closest comparison since it also uses random exploration, leads IRIS by $2.6$ pp on the simpler Spatial tasks, but it grounds subgoals by side-view pose estimation and falls $23.3$ pp behind on the occlusion- and contact-heavy LIBERO-Goal, as on the real robot (Section~\ref{subsec:real_robot_eih}).} Among methods reporting all four suites, IRIS's $89.3\%$ average exceeds SSI-Policy, the next-best method, by $8.9$ pp. Like IRIS, ATM and SSI-Policy train their video models on the action-free videos of all demonstrations, whereas UniVLA and OpenVLA-OFT rely on external pretrained models.}
}

\begin{table}[t]
    \vspace*{0.25mm}
    \centering
    \caption{\footnotesize \textbf{Low-data performance comparison on LIBERO using $20\%$ of task-specific demonstrations (success rate, \%).} $^{\dagger}$: video model trained on the action-free videos of all demonstrations; {$^{*}$: initialized from external pretrained models}; $^{\ddagger}$: no action-labeled demonstrations, side view only; --: unavailable.}
    \label{tab:data_efficiency}
    \vspace{-1mm}
    \scriptsize
    \setlength{\tabcolsep}{4pt}
    \renewcommand{\arraystretch}{1.05}
    \begin{tabular*}{\columnwidth}{@{\extracolsep{\fill}}lccccc@{}}
    \toprule
    \textbf{Method} & \textbf{Spatial} & \textbf{Object} & \textbf{Goal} & \textbf{Long} & \textbf{Average} \\
    \midrule
    Diffusion Policy~\cite{diffusion_policy} & 67.7 & 78.0 & 35.0 & 37.3 & 54.5 \\
    MaIL~\cite{MaIL_Jia_2024_CoRL} & 60.9 & 85.4& 56.3 & 46.0 & 62.2 \\
    ATM$^{\dagger}$~\cite{atm}               & 68.5 & 68.0 & 77.8 & 39.3 & 63.4 \\
    UniVLA$^{*}$~\cite{univla}               & --   & --   & \textbf{90.4} & \underline{71.4} & --   \\
    SSI-Policy$^{\dagger}$~\cite{ssi-policy} & 83.5 & \underline{95.0} & 82.5 & 60.7 & \underline{80.4} \\
    OpenVLA-OFT$^{*}$~\cite{Openvla-oft}     & --   & --   & 88.2 & 62.2 & --   \\
    GVF-TAPE$^{\ddagger}$~\cite{gvf-tape}    & \textbf{95.5} & 86.7 & 66.8 & -- & -- \\
    \rowcolor{gray!10}[0pt][0pt]
    \textbf{IRIS (Ours)}$^{\dagger}$          & \underline{92.9} & \textbf{97.6} & \underline{90.1} & \textbf{76.7} & \textbf{89.3} \\
    \rowcolor{gray!10}[0pt][0pt]
    \multicolumn{1}{r}{\tiny\color{gray}std over seeds} & {\tiny\color{gray}$\pm$1.7} & {\tiny\color{gray}$\pm$0.3} & {\tiny\color{gray}$\pm$3.1} & {\tiny\color{gray}$\pm$5.6} & {\tiny\color{gray}$\pm$1.4} \\
    \bottomrule
    \end{tabular*}
    \vspace{-0.25mm}
    \end{table}

\subsection{Demonstration Efficiency of Exploration Pretraining}
\label{subsec:rand_data}

\begin{figure}[t]
        \centering        \includegraphics[width=0.75\columnwidth]{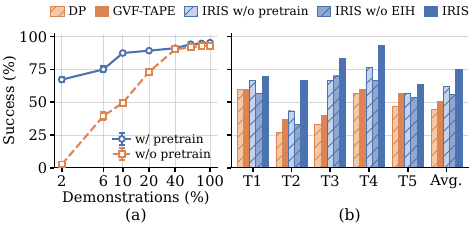}
        \vspace{-3mm}
    \caption{\footnotesize \textbf{Ablations of random-exploration pretraining and eye-in-hand conditioning.} (a) LIBERO success versus the fraction of demonstrations used to fine-tune the IDM, with and without pretraining. (b) Real-world success over $30$ trials per task for IRIS, its two ablations, Diffusion Policy (DP), and GVF-TAPE; T1--T5 as in Fig.~\ref{fig:real_world_setup}, Avg.\ over all $150$ trials. EIH: eye-in-hand.}
    \label{fig:ablation}
    \end{figure}
    
{With the video planner fixed, we vary only the demonstrations used to fine-tune the IDM. Figure~\ref{fig:ablation}(a) shows that pretraining improves success by $64.7$ and $38.0$ pp at the $2\%$ and $10\%$ budgets, respectively, but by only $2.7$ pp with the full demonstration set. Exploration pretraining therefore reduces dependence on task-specific demonstrations rather than providing a uniform gain; this is the source of the low-data robustness in Table~\ref{tab:data_efficiency}. {It also stabilizes fine-tuning: above a $40\%$ budget, the seed spread is $0.3$--$0.5$ pp with pretraining versus $1.9$--$2.6$ pp without.}}

\begin{figure}[t]
    \centering
    \includegraphics[width=0.75\columnwidth]{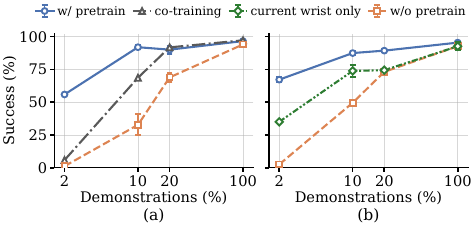}
    \vspace{-3mm}
    \caption{\footnotesize \textbf{Ablations of the IDM training scheme.} (a) Co-training versus two-stage training (\emph{w/ pretrain}) with matched data and total training samples; co-training uses a single seed. (b) Pretraining without the eye-in-hand subgoal; in \emph{current wrist only}, the IDM sees only the current eye-in-hand image in both training stages.}
    \label{fig:cotrain}
    \end{figure}

{To disentangle the contribution of the exploration data from that of the two-stage training scheme, we compare IRIS on LIBERO-Goal with a co-training baseline that trains the IDM in a single stage on the exploration and demonstration data, sampled $1{:}1$, with the same total number of training samples and a single seed. Figure~\ref{fig:cotrain}(a) shows that co-training matches IRIS at the $20\%$ and full budgets, {consistent with the gain there coming from the exploration data itself}; at $10\%$ and $2\%$ it falls $23.3$ and $50.1$ pp behind. Two-stage training thus matters most when demonstrations are scarce, and it requires no mixing ratio: one pretrained IDM serves every demonstration budget.}

\subsection{Controlled Ablation of the IDM-Side Components}
\label{subsec:component_ablation}

\begin{table}[t]
\vspace*{0.25mm}
\centering
\caption{\footnotesize \textbf{Crossed ablation of random-exploration pretraining and eye-in-hand conditioning on LIBERO-Goal (success rate, \%).} Gains in percentage points.}
\label{tab:crossed_ablation}
\vspace{-1mm}
\scriptsize
\setlength{\tabcolsep}{4pt}
\renewcommand{\arraystretch}{1.1}
\resizebox{1.0\columnwidth}{!}{%
\begin{tabular*}{\columnwidth}{@{\extracolsep{\fill}}lccc@{}}
\toprule
 & \textbf{W/ pretrain} & \textbf{W/o pretrain} & \textbf{Pretrain gain} \\
\midrule
W/ eye-in-hand   & $96.6 \pm 0.7$ & $94.1 \pm 1.2$ & $+2.5$ \\
W/o eye-in-hand  & $92.1 \pm 0.8$ & $91.3 \pm 3.9$ & $+0.8$ \\
\midrule
Eye-in-hand gain & $+4.5$ & $+2.8$ & \\
\bottomrule
\end{tabular*}}
\vspace{-0.25mm}
\end{table}

{Table~\ref{tab:crossed_ablation} {crosses} random-exploration pretraining and eye-in-hand conditioning on LIBERO-Goal under the full demonstration budget. {The four variants} share the same architecture, task demonstrations, and generated side-view subgoals; the variant with neither is a goal-conditioned Diffusion Policy IDM. Eye-in-hand conditioning improves success by $4.5$ pp with pretraining and $2.8$ pp without, while pretraining adds $2.5$ {and $0.8$ pp, respectively; the latter} lies within the seed spread of the no-eye-in-hand variants. Together with Fig.~\ref{fig:ablation}(a), this suggests complementary roles: eye-in-hand observations provide action-relevant local {information}, while {pretraining matters most as demonstrations become scarce}.}

{To isolate the future eye-in-hand subgoal used in Stage 1, we train a variant without conditioning on that input while retaining the current eye-in-hand image and all side-view inputs; Stage 2 and all other settings match IRIS. Averaged over the four suites, it trails IRIS by $32.2$, $13.6$, and $14.9$ pp at the $2\%$, $10\%$, and $20\%$ demonstration budgets, respectively, and by $2.8$ pp with all demonstrations (Fig.~\ref{fig:cotrain}(b)). Thus, future eye-in-hand conditioning during exploration pretraining contributes most when demonstrations are scarce.}

\subsection{Cross-Suite Generalization of Exploration Pretraining}
\label{subsec:transfer}

\begin{figure}[t]
    \centering
    \includegraphics[width=0.825\columnwidth]{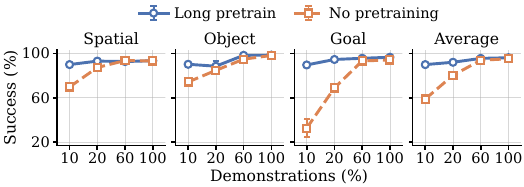}%
    \vspace{-1mm}
    \caption{\footnotesize \textbf{Transfer of LIBERO-Long exploration pretraining to other suites.} {\emph{Long pretrain} versus \emph{no pretraining}, each fine-tuned on the target suite. \emph{Average}: mean over the three target suites.}}
    \label{fig:transfer}
    \end{figure}

\begin{table}[t]
\vspace*{0.25mm}
\centering
\caption{\footnotesize {\textbf{Per-task effect of eye-in-hand conditioning under a data-stressed IDM} (IDM fine-tuned on $20\%$ of demonstrations per suite; success rate, \%). EIH: eye-in-hand.}}
\label{tab:eih_cases}
\vspace{-1mm}
\scriptsize
\setlength{\tabcolsep}{3pt}
\renewcommand{\arraystretch}{1.1}
\begin{tabular*}{\columnwidth}{@{\extracolsep{\fill}}llccc@{}}
\toprule
\textbf{Task} & \textbf{Dominant property} & \textbf{W/ EIH} & \textbf{W/o EIH} & \textbf{Gain} \\
\midrule
Goal T4     & Precision, contact, occlusion    & 90 & 66 & $+24$ \\
Long T5     & Precision, contact, occlusion    & 38 & \phantom{0}0 & $+38$ \\
Goal T6     & Contact, precision               & 98 & 46 & $+52$ \\
Long T7     & Contact, precision               & 24 & \phantom{0}0 & $+24$ \\
Long T9     & Precision                        & 96 & 78 & $+18$ \\
Spatial T5  & Precision                        & 98 & 82 & $+16$ \\
Spatial T10 & Precision                        & 96 & 84 & $+12$ \\
Object T6   & Precision                        & 96 & 84 & $+12$ \\
\bottomrule
\end{tabular*}
\vspace{-0.25mm}
\end{table}

{To test whether exploration data must be recollected for each new scene,} we pretrain the IDM solely on LIBERO-Long exploration and fine-tune it separately on each of the other three suites, against an otherwise identical model without pretraining (Fig.~\ref{fig:transfer}). The pretrained variant yields a higher three-suite average at the low budgets, by $+11.7$ pp at $20\%$ and $+31.0$ pp at $10\%$ of demonstrations, suggesting that exploration collected in one suite provides a reusable initialization whose benefit concentrates where target-suite demonstrations are scarce.

\subsection{Effect of Eye-in-Hand Conditioning}
\label{subsec:real_robot_eih}

{We evaluate whether the proposed eye-in-hand conditioning improves execution when task-relevant state is incomplete or ambiguous from the side view. In both real-world and simulation experiments, we compare full IRIS with an otherwise identical variant that removes the eye-in-hand input; the planner, demonstrations, and random-exploration pretraining are kept fixed.}

{\textbf{Real-world evaluation.}
Across the five tasks, IRIS succeeds in $75.3\%$ of trials, against $44.7\%$ for Diffusion Policy and $50.7\%$ for GVF-TAPE trained on the same demonstrations (Fig.~\ref{fig:ablation}(b)). Removing eye-in-hand conditioning lowers the average to $56.0\%$ and removing exploration pretraining to $62.0\%$, so both IDM-side interventions also help on the real robot. The eye-in-hand gain is largest on T2 ($+33$ pp) and T4 ($+27$ pp), the two occlusion tasks, where the plate and the microwave door are often hidden from the side view but visible from the wrist camera. {The pretraining gain is $17$--$23$ pp on T2--T4; T1 and T5 differ by only one or two successes out of $30$.
}

{\textbf{LIBERO task-level evaluation.}
We further examine eight LIBERO tasks, spanning all four suites, whose execution depends on local visual feedback that the side view does not reliably provide; Table~\ref{tab:eih_cases} lists them with the property that dominates each. Reducing the demonstration budget makes the contribution of the wrist observation easier to isolate. The gains range from $+12$ to $+52$ pp, with the largest improvement on pushing a plate to the front of the stove (Goal~T6). Two long-horizon tasks fall to $0\%$ without the wrist view, indicating that eye-in-hand feedback can be critical for some precision-and-contact {tasks}.}

\subsection{Alternative Eye-in-Hand Integration Strategies}
\label{subsec:eih_integration}

{On LIBERO-Long, we compare IRIS with two alternative uses of the wrist view (Fig.~\ref{fig:design}(a)). Dual-view generation predicts both side-view and wrist-view subgoals, following LingBot-VA and Vidar~\cite{lingbot-va,vidar}; residual refinement keeps a single-view IDM and uses the live wrist observation to predict a residual action correction.} IRIS outperforms these alternatives by $20.8$ and $10.0$ pp, respectively. {Because the wrist viewpoint moves with the robot, generated wrist subgoals can differ from the ground-truth frames used in IDM training; residual refinement avoids this mismatch but incorporates the wrist observation only after the nominal action has been inferred.} IRIS instead conditions the IDM directly on the live wrist observation while generating only side-view subgoals, {and masks} the eye-in-hand subgoal during fine-tuning (Section~\ref{subsec:idm_training}) {to match deployment}.}

\subsection{Sensitivity to Subgoal Stride and Quality}
\label{subsec:stride}

{The subgoal stride $m$ controls both the interval between subgoals and the motion the IDM must execute between them. On LIBERO-Goal, success varies by only $2.9$ pp for $m\in\{3,5,8\}$, but falls to $33.1\%$ at $m=1$ (Fig.~\ref{fig:design}(b)). Near-identical consecutive subgoals may provide weak motion cues, while more subgoals per rollout create more opportunities for errors to accumulate.}

{To test sensitivity to inaccurate generated subgoals, we add Gaussian noise ($\sigma\in\{0.05,0.1,0.2,0.3\}$ for pixels in $[0,1]$) or Gaussian blur ($1$--$2$ px) before IDM inference on LIBERO-Goal (single seed). Noise through $\sigma=0.2$ reduces success by at most $7.0$ pp, but performance drops sharply at $\sigma=0.3$; a $2$ px blur reduces success by $9.2$ pp despite having higher PSNR than the $\sigma=0.2$ noise (Fig.~\ref{fig:design}(c)). Thus, PSNR alone does not predict downstream success. These tests cover appearance corruption, not incorrect object motion.}

\begin{figure}[t]
\vspace*{0.25mm}
\centering
\includegraphics[width=0.75\columnwidth]{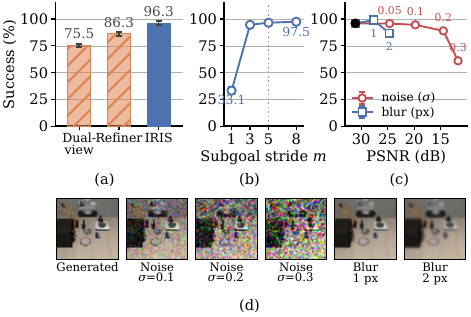}
\vspace{-2mm}
\caption{\footnotesize \textbf{Eye-in-hand integration alternatives and sensitivity to subgoal stride and quality.} {(a) Dual-view generation and residual refinement on LIBERO-Long. (b) LIBERO-Goal success versus stride $m$; dotted line: default $m=5$. (c) LIBERO-Goal success with noise- or blur-corrupted subgoals versus PSNR to true future frames; black dot: uncorrupted planner. (d) Example corruptions. Error bars in (a)--(b): one standard deviation.}}
\label{fig:design}
\vspace{-0.25mm}
\end{figure}

%% file: figs/realworld_setup.tex
\begin{figure*}[t]
\vspace*{0.25mm}
\centering
\begin{minipage}[t]{0.2180\textwidth}\centering\includegraphics[width=\linewidth]{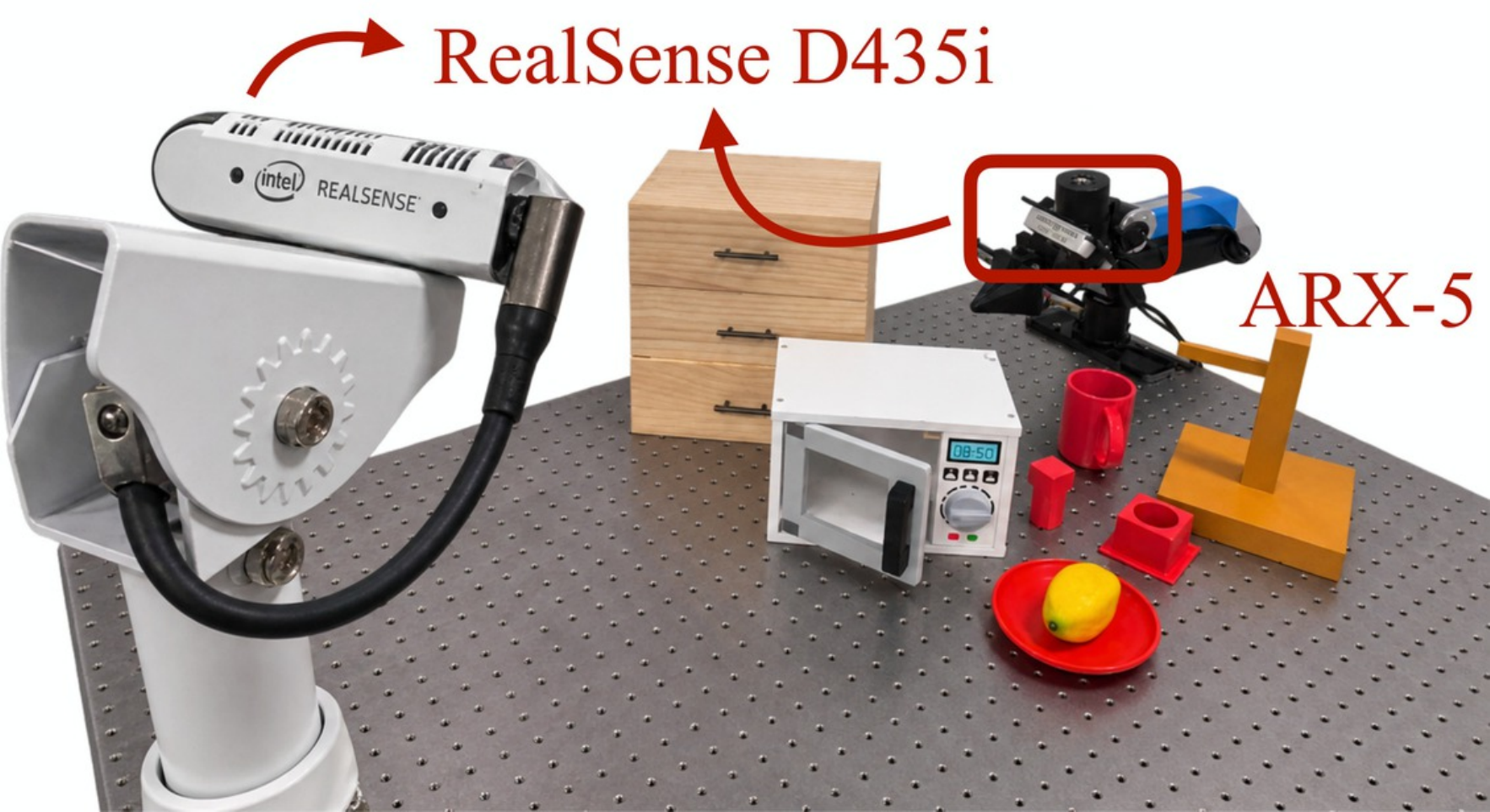}\par\vspace{-6pt}{\footnotesize\bfseries Setup\strut}\end{minipage}\hfill
\begin{minipage}[t]{0.1474\textwidth}\centering\begin{tikzpicture}[inner sep=0]\node[anchor=south west] (im) {\includegraphics[width=\linewidth]{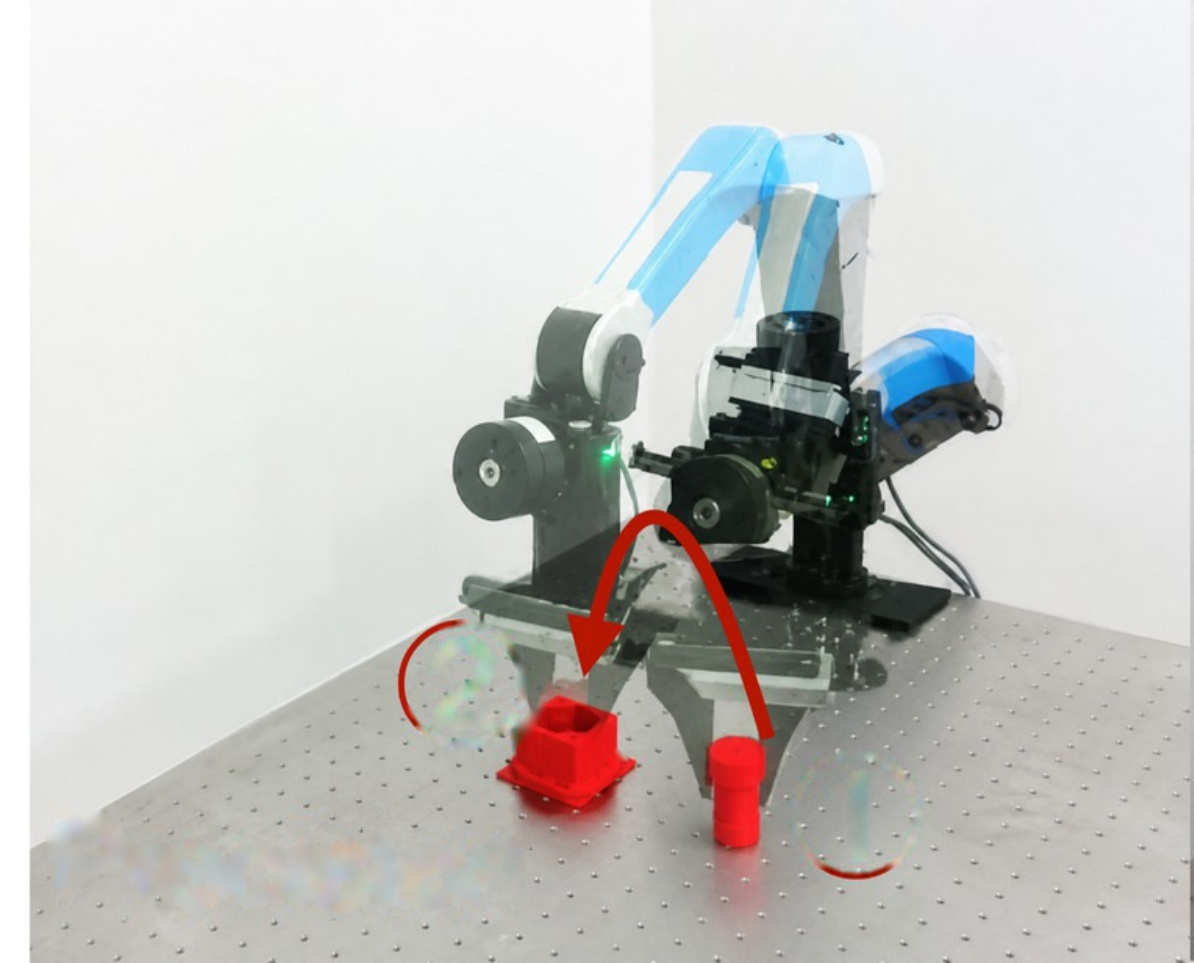}};\node[anchor=south west,xshift=2pt,yshift=2pt,fill=white,fill opacity=0.82,text opacity=1,rounded corners=1pt,inner sep=1.4pt,font=\scriptsize] at (im.south west) {Precision};\node[circle,draw=red!85!black,line width=0.5pt,minimum size=7.5pt,inner sep=0pt,font=\tiny,text=red!85!black] at ([xshift=0.7170\linewidth,yshift=0.1363\linewidth]im.south west) {1};\node[circle,draw=red!85!black,line width=0.5pt,minimum size=7.5pt,inner sep=0pt,font=\tiny,text=red!85!black] at ([xshift=0.3990\linewidth,yshift=0.2290\linewidth]im.south west) {2};\end{tikzpicture}\par\vspace{-6pt}{\footnotesize\bfseries T1\strut}\end{minipage}\hfill
\begin{minipage}[t]{0.1429\textwidth}\centering\begin{tikzpicture}[inner sep=0]\node[anchor=south west] (im) {\includegraphics[width=\linewidth]{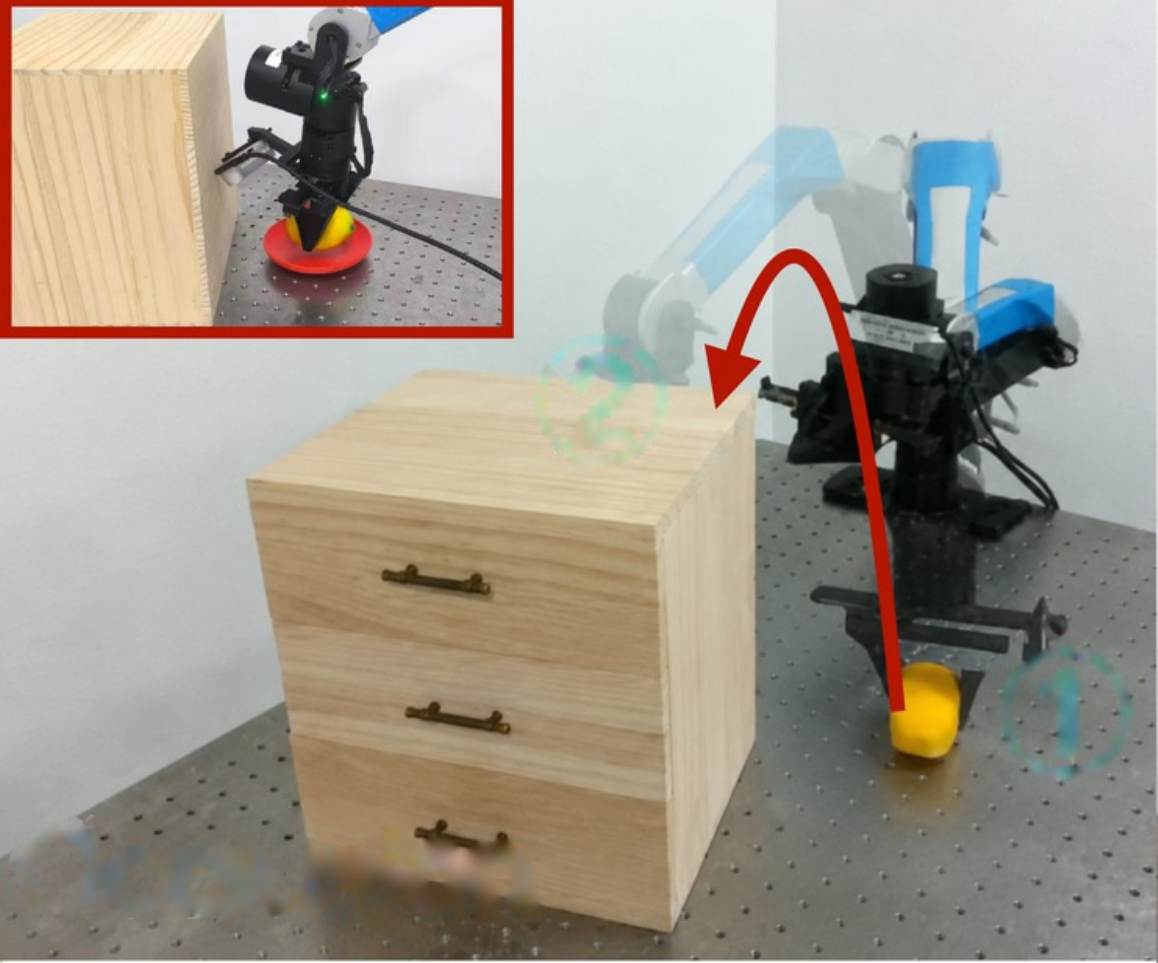}};\node[anchor=south west,xshift=2pt,yshift=2pt,fill=white,fill opacity=0.82,text opacity=1,rounded corners=1pt,inner sep=1.4pt,font=\scriptsize] at (im.south west) {Occlusion};\node[circle,draw=red!85!black,line width=0.5pt,minimum size=7.5pt,inner sep=0pt,font=\tiny,text=red!85!black] at ([xshift=0.9157\linewidth,yshift=0.2305\linewidth]im.south west) {1};\node[circle,draw=red!85!black,line width=0.5pt,minimum size=7.5pt,inner sep=0pt,font=\tiny,text=red!85!black] at ([xshift=0.5195\linewidth,yshift=0.4845\linewidth]im.south west) {2};\end{tikzpicture}\par\vspace{-6pt}{\footnotesize\bfseries T2\strut}\end{minipage}\hfill
\begin{minipage}[t]{0.1437\textwidth}\centering\begin{tikzpicture}[inner sep=0]\node[anchor=south west] (im) {\includegraphics[width=\linewidth]{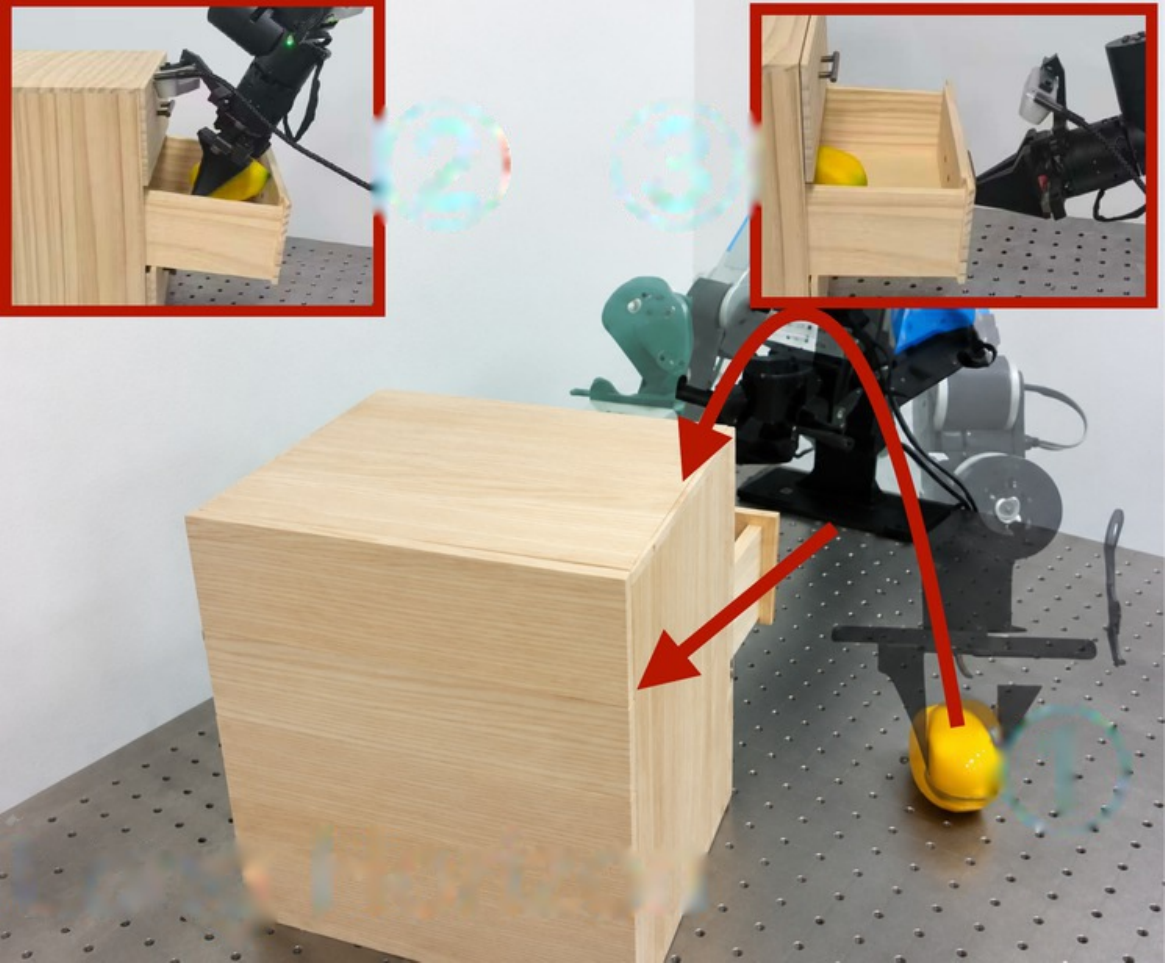}};\node[anchor=south west,xshift=2pt,yshift=2pt,fill=white,fill opacity=0.82,text opacity=1,rounded corners=1pt,inner sep=1.4pt,font=\scriptsize] at (im.south west) {Long horizon};\node[circle,draw=red!85!black,line width=0.5pt,minimum size=7.5pt,inner sep=0pt,font=\tiny,text=red!85!black] at ([xshift=0.9064\linewidth,yshift=0.1656\linewidth]im.south west) {1};\node[circle,draw=red!85!black,line width=0.5pt,minimum size=7.5pt,inner sep=0pt,font=\tiny,text=red!85!black] at ([xshift=0.3705\linewidth,yshift=0.6852\linewidth]im.south west) {2};\node[circle,draw=red!85!black,line width=0.5pt,minimum size=7.5pt,inner sep=0pt,font=\tiny,text=red!85!black] at ([xshift=0.5925\linewidth,yshift=0.6859\linewidth]im.south west) {3};\end{tikzpicture}\par\vspace{-6pt}{\footnotesize\bfseries T3\strut}\end{minipage}\hfill
\begin{minipage}[t]{0.1601\textwidth}\centering\begin{tikzpicture}[inner sep=0]\node[anchor=south west] (im) {\includegraphics[width=\linewidth]{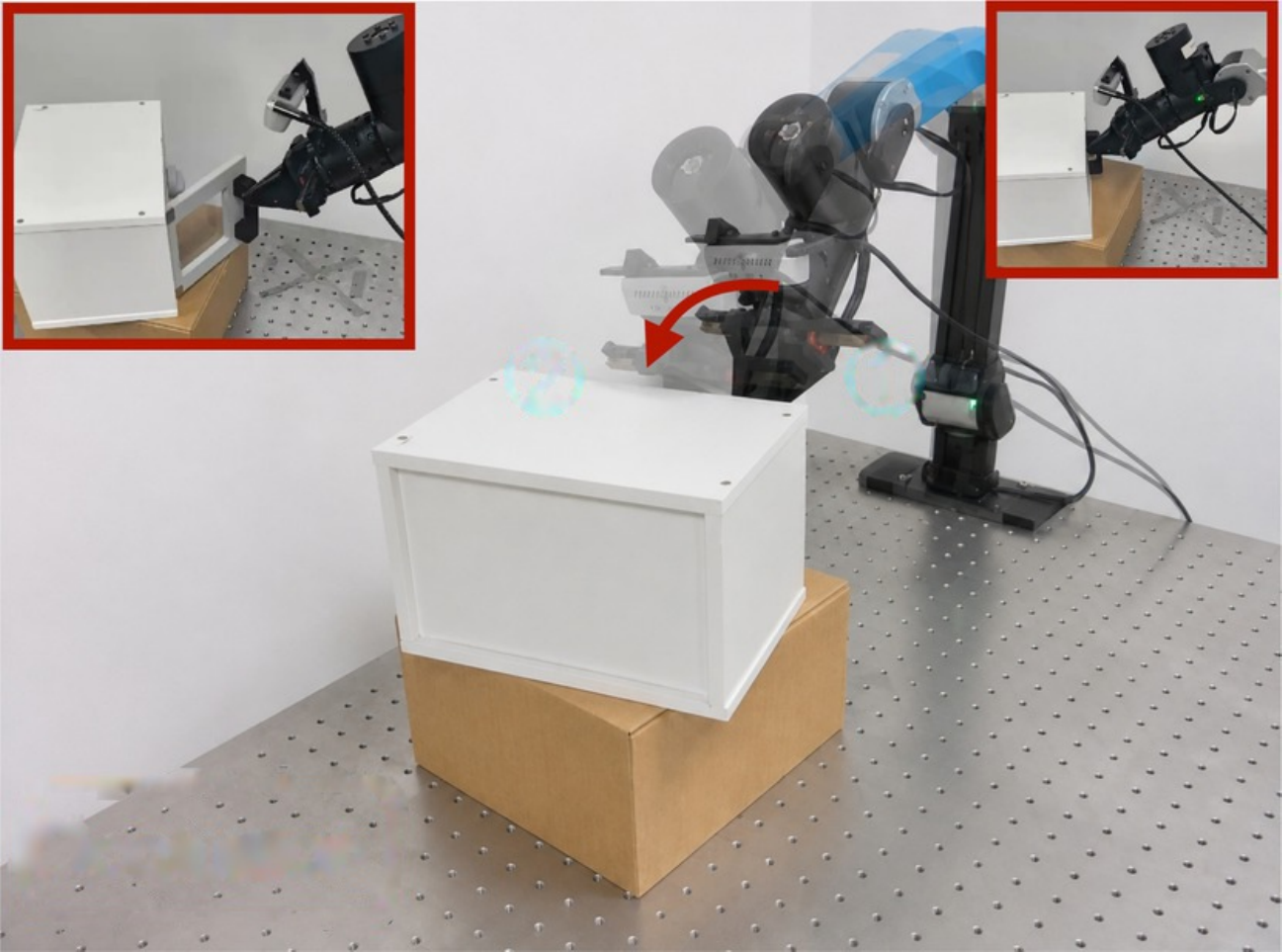}};\node[anchor=south west,xshift=2pt,yshift=2pt,fill=white,fill opacity=0.82,text opacity=1,rounded corners=1pt,inner sep=1.4pt,font=\scriptsize] at (im.south west) {Occlusion};\node[circle,draw=red!85!black,line width=0.5pt,minimum size=7.5pt,inner sep=0pt,font=\tiny,text=red!85!black] at ([xshift=0.6894\linewidth,yshift=0.4484\linewidth]im.south west) {1};\node[circle,draw=red!85!black,line width=0.5pt,minimum size=7.5pt,inner sep=0pt,font=\tiny,text=red!85!black] at ([xshift=0.4245\linewidth,yshift=0.4489\linewidth]im.south west) {2};\end{tikzpicture}\par\vspace{-6pt}{\footnotesize\bfseries T4\strut}\end{minipage}\hfill
\begin{minipage}[t]{0.1580\textwidth}\centering\begin{tikzpicture}[inner sep=0]\node[anchor=south west] (im) {\includegraphics[width=\linewidth]{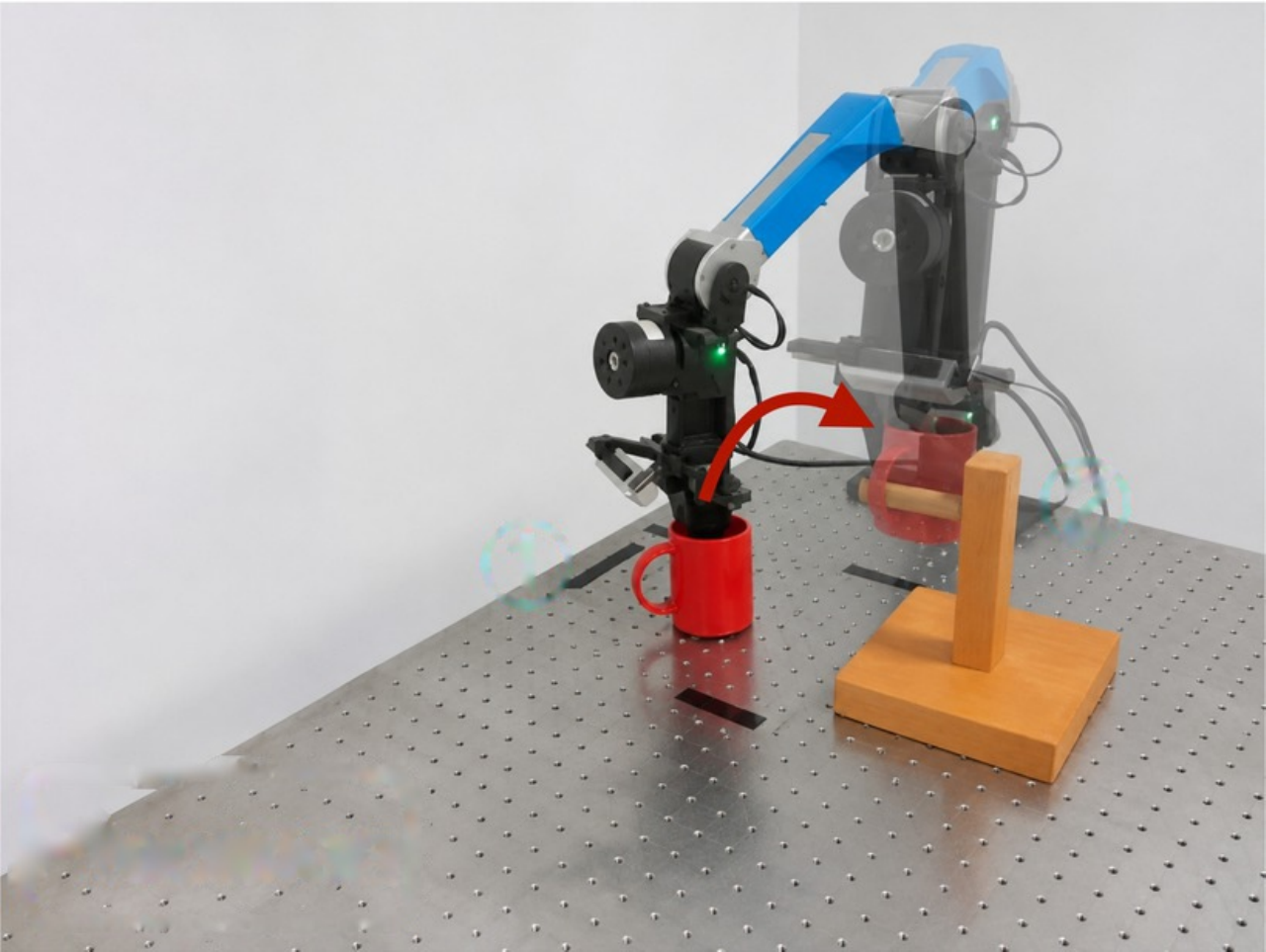}};\node[anchor=south west,xshift=2pt,yshift=2pt,fill=white,fill opacity=0.82,text opacity=1,rounded corners=1pt,inner sep=1.4pt,font=\scriptsize] at (im.south west) {Precision};\node[circle,draw=red!85!black,line width=0.5pt,minimum size=7.5pt,inner sep=0pt,font=\tiny,text=red!85!black] at ([xshift=0.4164\linewidth,yshift=0.3069\linewidth]im.south west) {1};\node[circle,draw=red!85!black,line width=0.5pt,minimum size=7.5pt,inner sep=0pt,font=\tiny,text=red!85!black] at ([xshift=0.8574\linewidth,yshift=0.3544\linewidth]im.south west) {2};\end{tikzpicture}\par\vspace{-6pt}{\footnotesize\bfseries T5\strut}\end{minipage}
\vspace{-2mm}
\caption{\footnotesize \textbf{{Real-world setup and tasks.}}
Left: ARX-5 with side-view and wrist cameras.
T1--T5: block insertion, lemon placement on a plate, lemon placement in a drawer and drawer closure, microwave-door closure, and mug hanging.
Numbers mark task steps.}
\label{fig:real_world_setup}
\vspace{-0.25mm}
\end{figure*}

%% file: sections/limitation_future_work.tex
\section{Limitations and Future Work}
\label{sec:limitations}

{Although IRIS improves action grounding without a pretrained policy or video backbone, it still requires task-specific demonstrations to train the video planner and adapt the IDM, which externally pretrained video or vision--language backbones could help reduce. We have not tested exploration pretraining on an IDM initialized from large-scale robot data, which we leave to future work. Random exploration is also unguided, so curiosity-driven or goal-diverse strategies could broaden coverage of rare contact states. Confining eye-in-hand sensing to the IDM keeps planning scalable, while tasks driven by contact or tactile cues would need additional sensing modalities, {which this interface accepts}.}

%% file: sections/conclusion.tex
\section{Conclusion}

We presented IRIS, a video-based manipulation framework that improves imagine-then-execute policies by strengthening the inverse dynamics model rather than requiring more complex visual planning. {The pilot study showed that action grounding can remain a major bottleneck even with oracle subgoal frames.} IRIS addresses this bottleneck with two IDM-side interventions: task-agnostic random-exploration pretraining to broaden action-grounding coverage, and eye-in-hand-aware two-stage training to provide local execution feedback without requiring eye-in-hand video generation. {Experiments show improved generalization, data efficiency, precision, and competitive performance against strong baselines.}

%% file: appendix/implementation.tex
{\noindent\textbf{Implementation details.}
The planner concatenates a time-repeated side-view RGB frame
{channel-wise} with noisy future RGB-D frames; its
201M-parameter 3D U-Net resamples spatially while preserving temporal
resolution. It uses a frozen CLIP ViT-B/32 text encoder
{whose hidden states pass through a two-layer resampler before
projection to the U-Net time-embedding dimension}, $H=6$ subgoals at
stride $m=5$, $128\times128$ resolution, batch size 8, 100k training
steps, and three inference steps. For LIBERO-Long, batch size and
training steps increase to 32 and 200k. A {100k-step} planner
run takes about 16\,h on two A100s. The 350M-parameter 1D U-Net IDM
uses a ResNet-18 encoder per visual input, seven-dimensional actions,
100 diffusion timesteps, and ten inference steps. Both models use AdamW {($10^{-4}$, cosine schedule)}. For
simulation/real-world training, the IDM uses batch sizes 128/256,
pretraining epochs 200/300, fine-tuning epochs 200/1000, and
resolutions $128\times128$/$240\times320$.}